\documentclass[10pt,twocolumn,letterpaper]{article}

\usepackage[letterpaper,margin=1in]{geometry}
\usepackage{times}
\usepackage[utf8]{inputenc}
\usepackage[T1]{fontenc}
\usepackage{csquotes}
\usepackage{graphicx}
\usepackage{booktabs}
\usepackage{array}
\usepackage{multirow}
\usepackage{amsmath}
\usepackage{amssymb}
\usepackage{xcolor}
\usepackage{xspace}
\usepackage{pifont}
\usepackage{tikz}
\usetikzlibrary{arrows.meta,positioning,fit,backgrounds,calc}
\usepackage{tcolorbox}
\usepackage{caption}
\usepackage{subcaption}
\usepackage{microtype}
\usepackage{enumitem}
\usepackage{natbib}
\usepackage{balance}
\setcitestyle{authoryear,round,citesep={;},aysep={,},yysep={;}}
\usepackage[colorlinks]{hyperref}
\usepackage[capitalize]{cleveref}

\definecolor{teacherblue}{RGB}{26,105,166}
\definecolor{voiceblue}{RGB}{52,128,205}
\definecolor{tracepurple}{RGB}{102,75,170}
\definecolor{evaluatorgreen}{RGB}{25,132,88}
\definecolor{gapred}{RGB}{186,50,55}
\definecolor{softgray}{RGB}{245,247,249}
\hypersetup{urlcolor=teacherblue,citecolor=teacherblue,linkcolor=teacherblue}

\newcommand{\benchname}{\textsc{ShowTellArena}\xspace}

\newcommand{\Paragraph}[1]{\vspace{0.4ex}\noindent\textbf{#1}\hspace{0.4em}}

\newcommand{\snapshotdate}{20 September 2026\xspace}

\newcommand{\numtested}{39\xspace}
\newcommand{\numshared}{28\xspace}
\newcommand{\numuntested}{3\xspace}
\newcommand{\numattempts}{218\xspace}

\newcommand{\brackettcomplete}{107\xspace}
\newcommand{\brackettattempts}{107\xspace}

\newcommand{\codexcomplete}{51\xspace}
\newcommand{\codexattempts}{51\xspace}

\newcommand{\claudecomplete}{12\xspace}
\newcommand{\claudeattempts}{60\xspace}
\newcommand{\reportedmodels}{Codex: model not recorded (51 attempts); Claude: Sonnet 5 (60 attempts)\xspace}

\newcommand{\brackettrepeatsummary}{three selected attempts on 29 cases and two on 10\xspace}

\newcommand{\codexrepeatsummary}{three selected attempts on 4 cases and two on 7 and one on 25\xspace}

\newcommand{\clauderepeatsummary}{two selected attempts on 30 cases\xspace}

\newcommand{\numexcluded}{16\xspace}

\newcommand{\brackettimported}{47\xspace}
\newcommand{\codeximported}{27\xspace}
\newcommand{\importedgrader}{Claude Sonnet 4.6\xspace}

\newcommand{\numreleased}{50\xspace}
\newcommand{\releasequestions}{502\xspace}
\newcommand{\releasenarration}{438\xspace}
\newcommand{\releasescreenshots}{7,182\xspace}
\newcommand{\releaseclosed}{253\xspace}
\newcommand{\releaserubric}{249\xspace}
\newcommand{\releaseapps}{5\xspace}

\setlist[itemize]{leftmargin=*,nosep}
\setlist[enumerate]{leftmargin=*,nosep}

\title{\benchname: Evaluating Business Workflow\\Understanding from Demonstrations}
\author{David Garg \quad Ritobrata Sarkar \quad Ehsan Azarnasab \quad Siddhartha Borah\\
Brackett Labs\\
{\texttt{\{david,rito,ehsan,sid\}@brackett.ai}}\\
{Correspondence: \texttt{brackettlabs@brackett.ai}}}
\date{}
\hypersetup{pdftitle={ShowTellArena: Evaluating Business Workflow Understanding from Demonstrations},pdfauthor={David Garg, Ritobrata Sarkar, Ehsan Azarnasab, Siddhartha Borah}}

\begin{document}
\raggedbottom

\maketitle

\begin{abstract}
We often teach a colleague by showing the work and explaining the decisions as we go. How can we check what an agent understood from the same lesson? We introduce \benchname, a benchmark protocol and public dataset for comprehension after narrated business demonstrations. The v1.0 release contains \numreleased business workflow tasks, with recordings, screenshots, narration, fixture seeds, and \releasequestions questions. Tasks span finance, hiring, procurement, customer decisions, inventory, and logistics. The protocol holds the business scenario and quiz fixed while allowing each product to capture the lesson through its own teaching interface. Questions test operational rules, boundaries, exceptions, and errors in proposed automations. We analyze \numattempts selected pilot attempts across \numtested workflow cases, including \numshared cases attempted by all three evaluated systems. These exploratory results expose both answer errors and failures to complete the teaching experience. We describe the release's verification gaps and the pilot's uneven coverage, exclusions, and grading provenance. The contribution is an inspectable dataset and assessment workflow that others can extend; the selected pilot is not a controlled product ranking.
\end{abstract}

\section{Introduction}
\label{sec:intro}
When someone new joins, we often ask a colleague to show them the work. They open the relevant records, point to a value, and explain why it matters. They also explain when the usual rule stops applying and who can approve an exception. Much of this knowledge lives in the work itself rather than in a complete written procedure.

Showing an agent a process does not establish that it understood it. The agent may reproduce the clicks while missing the rule, remember the rule but miss an exception, or give a plausible answer that is wrong for this business. We want to check its understanding before asking it to handle the next case.

\benchname asks: \emph{after a narrated demonstration, what can the agent explain and apply?} Each task combines a repeatable application state, demonstrated actions and narration, and a fixed comprehension quiz. Products encounter the same intended lesson through their native teaching interfaces. Their capture, processing, and permitted learned state remain part of the system under evaluation.

This paper contributes (1) a task format linking questions to teaching evidence, (2) a public dataset of \numreleased workflow tasks and a harness for native Teach--Comprehend evaluation, and (3) an exploratory analysis of selected pilot attempts. All \numtested evaluated case names map to released tasks, although that mapping alone does not establish which recording and quiz revision each historical run used. The \href{https://github.com/bracketthq/showAndTell-arena}{code}, \href{https://huggingface.co/datasets/brackettai/showtellarena/tree/v1.0}{dataset}, and \href{https://bracketthq.github.io/showAndTell-arena/}{research site} are public.

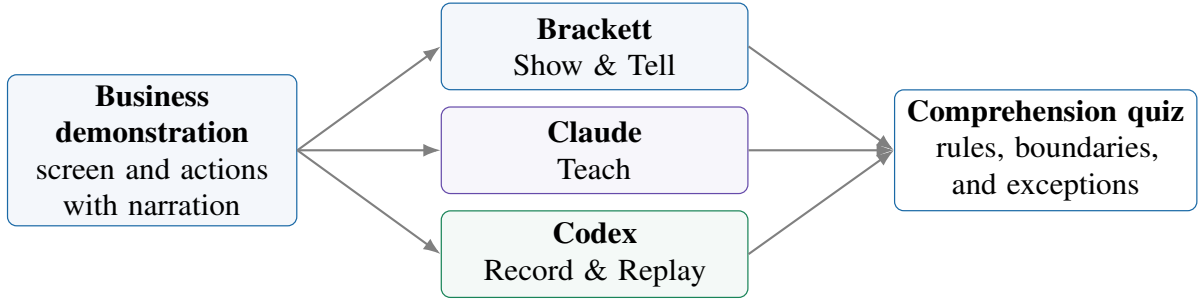
\begin{figure*}[t]
\centering
\resizebox{0.96\textwidth}{!}{\begin{tikzpicture}[
 box/.style={draw,rounded corners=3pt,minimum height=1.0cm,align=center,font=\normalsize},
 arrow/.style={-{Latex[length=2mm]},line width=0.7pt},
 note/.style={font=\normalsize,align=center}
]
\node[box,draw=teacherblue,fill=teacherblue!5,text width=3.1cm] (demo) at (0,0)
 {\textbf{Business demonstration}\\screen and actions\\with narration};
\node[box,draw=teacherblue,fill=teacherblue!5,text width=3.3cm] (brackett) at (5.1,1.2)
 {\textbf{Brackett}\\Show \& Tell};
\node[box,draw=tracepurple,fill=tracepurple!5,text width=3.3cm] (claude) at (5.1,0)
 {\textbf{Claude}\\Teach};
\node[box,draw=evaluatorgreen,fill=evaluatorgreen!5,text width=3.3cm] (codex) at (5.1,-1.2)
 {\textbf{Codex}\\Record \& Replay};
\node[box,draw=teacherblue,text width=3.25cm] (quiz) at (10.3,0)
 {\textbf{Comprehension quiz}\\rules, boundaries,\\and exceptions};
\foreach \product in {brackett,claude,codex}{
 \draw[arrow,gray] (demo.east) -- (\product.west);
 \draw[arrow,gray] (\product.east) -- (quiz.west);
}
\node[note] at (5.1,-2.0) {Each product captures and uses its own learned state.};
\end{tikzpicture}}
\caption{\textbf{A common lesson and quiz, native teaching interfaces.} Each product captures the demonstrated work and uses its own permitted learned state. The comparison includes capture and memory; it does not isolate foundation-model capability. This diagram describes the protocol, not proof of identical historical run conditions.}
\label{fig:teaser}
\end{figure*}

\section{Related Work}
\label{sec:related}
\Paragraph{Demonstration understanding.}
WONDERBREAD evaluates business-process documentation, knowledge transfer, and process improvement from demonstrations~\citep{wornow2024wonderbread}. VideoWebArena includes factual question answering and skill retention from video tutorials~\citep{jang2024videowebarena}. These are direct precedents: neither comprehension questions nor learning from demonstrations is new on its own. Our focus is evidence-linked business rules and exceptions assessed across products' native teaching interfaces.

\Paragraph{Executing business work.}
WorkArena and OSWorld evaluate agents acting in enterprise applications and computer environments~\citep{drouin2024workarena,xie2024osworld}. $\tau$-bench evaluates tool use and policy following through final state and repeated trials~\citep{yao2024taubench}. TheAgentCompany evaluates consequential work within a simulated company~\citep{xu2024agentcompany}. Their execution outcomes complement our immediate comprehension diagnostic. Correctly explaining a refund rule does not establish that the agent can carry out the refund safely.

\Paragraph{Learning across tasks.}
ApprenticeBench studies agents learning an accounts-payable job with historical work, instructions, and feedback across a sequence of tasks~\citep{neocognition2026apprentice}. Our present measurement stops after teaching and questioning. It does not establish learning gain, transfer to executed work, or delayed retention. Those require additional controls and observations.

\section{Task Format and Protocol}
\label{sec:benchmark}
A task has five parts: an initial application state, a demonstration, narration, a quiz, and a grading contract. We distinguish the \emph{teacher/evaluator package}, which contains seeds and expected answers, from the \emph{learner's permitted observations}. Publishing an answer key makes task design inspectable; it does not authorize access to that key during evaluation.

\subsection{Native Teach--Comprehend}
\label{sec:protocol}
The released harness provides adapters for Brackett Show \& Tell, Claude's teaching flow, and Codex Record \& Replay. It seeds the fixture applications, starts the product's recording interface, performs the demonstrated actions with narration, waits for processing, and asks the fixed questions. The repository documents authenticated product accounts and extension/plugin prerequisites; adapter availability is not a guarantee of access to every commercial product. Codex recording requires macOS. Browser adapters also document Linux support.

The controlled input is the intended scenario and scripted lesson, not an identical internal representation. Processing delays, capture failures, or omitted audio can change what the product actually receives. A comparison therefore needs a record of the completed demonstration, captured lesson, processing outcome, and permitted memory. In a prospective run, the task state and learner context must be reset according to the declared policy before every repetition.

At quiz time, the product may use the learned state and notes allowed by that policy. It must not read the evaluator's questions with answer fields, demonstration source, or fixture records to recover an answer. These paths need technical access controls and an audit of the actual tools available to the learner. A prompt asking it not to look is not an access-control mechanism.

\subsection{Questions and grading}
\label{sec:questions}
The task schema supports multiple-choice, closed-answer, and rubric questions. The returns example uses eight closed-answer questions and three rubric questions. Closed answers first undergo whole-answer normalization against accepted aliases; unmatched answers can be assessed by a configured semantic-equivalence judge. Rubric questions allow bounded partial credit. The current runner averages item grades and records the judge configuration and overrides. Its default model-assisted judge configuration is separate from the historical pilot's grading provenance.

An unavailable grader leaves the current runner's result ungraded. This differs from a learner that produces an incorrect answer and from a teaching flow that does not complete. The pilot workbook uses its own incomplete-attempt convention, described in Section~\ref{sec:scoring}; we do not retroactively present it as the released runner's canonical evaluation.

\subsection{Evidence-linked answers}
Each expected claim should be supported by what the learner was shown or told. A question can cite a recording interval, a screenshot, or a source record used to construct the scenario. Recording and screenshot references connect the answer to observed teaching. Source records can verify dates and amounts but do not prove that a detail appeared in the lesson. A reviewer must check answerability from the permitted observations, including which information is merely pointed to and which is actually visible or spoken.

\section{Public Dataset}
\label{sec:dataset}
The public Hugging Face v1.0 release contains \numreleased business workflow tasks at revision \texttt{e8dc213}. We pin it separately from code revision \texttt{ad59046}; full identifiers are in the accompanying release manifest. Tasks cover finance, hiring, procurement, customer decisions, inventory, and logistics across ERPNext, Fleetbase, ONLYOFFICE, Roundcube, and Twenty. Numbered variants are separate tasks, not necessarily independent workflow families.

\begin{table}[t]
\centering\normalsize
\begin{tabular}{@{}lr@{}}
\toprule
\textbf{Released material} & \textbf{Count} \\
\midrule
Workflow tasks / recordings & \numreleased / \numreleased \\
Application integrations & \releaseapps \\
Narration segments & \releasenarration \\
Screenshots & \releasescreenshots \\
Closed-answer / rubric questions & \releaseclosed / \releaserubric \\
Total questions & \releasequestions \\
\bottomrule
\end{tabular}
\caption{\textbf{Public dataset v1.0.} Counts are verified against the released bundles, not only the index. Screenshots and questions are observations within tasks, not independently sampled workflows. The evaluation covers a subset of this release.}
\label{tab:release}
\end{table}

\subsection{One inbox, different decisions}
Customer Returns Inbox Triage illustrates the task format with an authored returns-inbox scenario. Normal returns qualify within 30 days of delivery. The handler may approve refunds up to \pounds50 directly; larger eligible refunds go to finance with supporting details. An item that arrived damaged overrides both ordinary gates in this scenario. These are task-specific rules, not a general returns policy.

The learner must identify the relevant messages, use delivery dates rather than email dates, apply inclusive limits, and handle the exception. A boundary question asks how to process a normal \pounds41 return exactly 30 days after delivery and a \pounds50 return delivered five days earlier. Both qualify for direct approval. A second question asks about a \pounds74 item delivered more than 30 days ago that arrived cracked: the damage exception changes the decision.

A rubric question goes further by presenting a proposed automation that applies plausible but incorrect rules. The answer must diagnose errors such as escalating exactly \pounds50 or rejecting the damaged item solely because it is too old. This separates recalling one demonstrated click sequence from explaining how the rule applies to another case. It remains a question-answering test, not an executed transfer experiment.

\subsection{Construction and evidence}
Each bundle includes demonstration/replay artifacts, narration, application seed material, and questions with accepted answers or rubrics. The release contains 857 question-evidence references. These links support review of whether an answer was taught.

The returns question file documents revisions after replay inspection, including clearer cadence, batch scope, and finance-destination wording. It also explicitly records that fresh attentive-human positive controls and blind question-only controls remain pending. We therefore do not claim demonstrated inter-annotator agreement, validated question difficulty, or measured dependence on the lesson.

A structural audit found all referenced evidence paths present. Of 857 evidence references, 773 supplied hashes agree with published asset identities (Hub LFS metadata for binary assets; downloaded bytes for other files), 16 disagree, and 68 omit a hash. All 218 seed-asset references resolve and their supplied hashes match. Hash agreement establishes file identity, not whether the lesson supports an answer.

Fifteen mismatches belong to the returns example; one belongs to Job Requisition Triage. The returns quiz also retains an address differing from its screenshot evidence. The accompanying corrected returns quiz updates its 15 hashes and accepts both displayed and replay-fixture address forms, with an explicit change record. Its 23 evidence references verify. This targeted correction is separate from upstream v1.0 and does not repair or validate the entire dataset, or justify reinterpreting historical scores.

\subsection{Availability and intended use}
The benchmark code is MIT-licensed. The dataset card declares CC BY 4.0 for authored benchmark content. Application-derived content and third-party software retain their applicable upstream terms and notices. Running the task also requires the documented fixture applications and product access. The published index does not pin a complete fixture-image manifest, so the release does not establish bit-for-bit environment reproduction.

The dataset supports inspecting task construction, exercising adapters, and studying comprehension across workflows. Its authored scenarios do not establish general business competence. Public answer keys make it unsuitable as a hidden test against future training contamination. New studies should declare what material a learner could access and keep evaluator records separate during the run.

\section{Writing and Administering the Exam}
\label{sec:quality}
Writing a good question requires knowing which detail changes the decision. During development, plausible agent-written questions sometimes tested generic business knowledge or omitted the exception that mattered. We treat this as a development observation, not a quantified annotation study.

A useful review starts with an attentive person answering from the lesson and explaining the evidence. A question-only control then tests whether the answer is already supplied by prior knowledge or wording cues. Closed answers reduce option-based guessing but do not establish demonstration dependence. The revised public sample has not yet completed these controls.

Native teaching also creates different routes to the answer. Pilot notes describe attempts to inspect local code or fixture information, empty teaching artifacts, refusals, and interrupted processing. These observations motivate explicit access policies and failure-stage records. The workbook does not establish that all historical attempts passed a complete leakage audit.

A prospective evaluation should freeze the task revision, system configuration, repeat schedule, stopping rule, and grading before collecting results. Capture, processing, answer generation, grading, and infrastructure failures should be reported separately. Retain unsuccessful attempts and human interventions so that the cost of reaching an answer and the selection of retries remain visible.

\section{Exploratory Pilot}
\label{sec:experiments}
The workbook frozen on \snapshotdate contains \numattempts selected attempts across \numtested tested case entries; \numuntested additional entries are untested. This is the snapshot date, not a common execution date. Cases span finance, hiring, procurement, customer decisions, inventory, and logistics. Named variants remain separate entries and are not assumed to be independent workflow families. All \numtested evaluated case names have corresponding tasks in v1.0. The three untested workbook entries are not in that release, which also includes 11 additional tasks. Name correspondence is not proof of historical artifact identity.

\subsection{Coverage and aggregation}
\label{sec:scoring}
Brackett has \brackettrepeatsummary; Claude has \clauderepeatsummary; Codex has \codexrepeatsummary. Let $q_{ajr}\in[0,1]$ denote a recorded quiz score for system $a$, case $j$, and selected attempt $r$. We average attempts within each case, then give each shared case equal weight:
\begin{equation}
 C_a=\frac{100}{|J|}\sum_{j\in J}\frac{1}{R_{aj}}\sum_{r=1}^{R_{aj}}q_{ajr}.
\end{equation}
$J$ contains the \numshared cases attempted by every system; $R_{aj}$ is the selected attempt count for that system and case. Our aggregation treats NA as an incomplete attempt contributing zero, a blank as untested and excluded, and numeric zero as a graded answer. We recompute from attempt cells rather than spreadsheet summary formulas, which can omit text NA and zero values.

\begin{table}[t]
\centering\normalsize
\setlength{\tabcolsep}{2.5pt}
\begin{tabular}{@{}lrrr@{}}
\toprule
\textbf{System} & \textbf{Score (\%)} & \textbf{Graded/attempted} & \textbf{Coverage} \\
\midrule
Brackett & 83.8 & 83/83 & 39 \\
Codex & 55.4 & 32/32 & 36 \\
Claude & 16.7 & 12/56 & 30 \\
\bottomrule
\end{tabular}

\caption{\textbf{Selected pilot attempts on \numshared shared cases.} Score and graded/attempted counts use the shared set. Coverage counts tested cases across the full workbook. ``Graded'' means a numeric quiz score, not successful execution. Uneven exclusions and mixed grading limit comparability.}
\label{tab:main-results}
\end{table}

\subsection{What the records support}
Brackett is reported as a fused/multiple-model system. The workbook reports evaluated-model labels separately: \reportedmodels. These labels do not establish exact model or adapter versions at each run. Historical grades mix manual and model-assisted assessments; the workbook identifies \brackettimported imported Brackett grades and \codeximported imported Codex grades from \importedgrader, but it does not supply a common, independently calibrated grading record for all systems.

The workbook retains \numexcluded excluded attempt rows separately. Their selection rules are not uniform: some incomplete Codex attempts are excluded, whereas selected Claude NA attempts enter the score as zero. Consequently, Table~\ref{tab:main-results} describes the workbook's selected sample, not an unbiased ranking. Appendix~\ref{app:selection} reports the exclusions and the limits of a sensitivity check.

Across all tested cases, Brackett has \brackettcomplete numeric grades in \brackettattempts selected attempts, Codex \codexcomplete in \codexattempts, and Claude \claudecomplete in \claudeattempts. Newer attempt notes also document demonstration and replay problems. A difference in score can reflect capture, processing, answer production, or grading as well as comprehension.

We do not infer a causal benefit of teaching: there is no matched cold condition for these runs. Nor do we report confidence intervals, delayed retention, execution success, or question-family scores from these selected aggregates. The public quiz has been revised and is not asserted to be the version used in every historical attempt. The row-level export permits reproduction of this aggregation; missing historical artifacts prevent a complete rerun or independent regrading of the pilot.

\section{Limitations and Conclusion}
\label{sec:conclusion}
\benchname asks what an agent can explain after a business demonstration through its own teaching interface. It contributes a task format, evidence-linked questions, native adapters, and \numreleased released workflow tasks. The pilot illustrates why both answers and failures to reach the quiz matter.

The tasks are authored scenarios with related variants, not a representative sample of all business work. Independent answerability review and question-only controls remain necessary. The pilot has selected repetitions, incomplete version and access records, and mixed grading. Its authors develop Brackett, one of the evaluated systems; independent replication is needed before interpreting differences as a product ranking.

Future studies can add workflows with verified evidence and predeclared repetitions, then examine execution, new cases, and retention separately. We invite contributors to try an adapter, contribute a workflow, or help write a better exam.

\Paragraph{Preparation disclosure.}
OpenAI Codex assisted manuscript editing, code preparation, and consistency checks. The submitting authors are responsible for the final claims, citations, and released material.

{\normalsize\raggedright
\bibliography{refs}
\bibliographystyle{iclr2024_conference}
}

\clearpage
\appendix
\balance
\section{Release and Reproduction}
\label{app:release}
The inspected code revision is \texttt{ad59046} (20 September 2026 inspection). The public dataset v1.0 resolves to \texttt{e8dc213}, inspected on 20 September 2026. Full revision identifiers and evidence-file hashes are recorded in the accompanying release manifest. A corrected quiz with separately recorded hash and fixture-address changes is clearly separated from that upstream revision; neither its existence nor this paper changes the upstream release.

The ancillary package includes a sanitized pilot row export and a small aggregation script. These reproduce the table's equal-case weighting without distributing the private workbook, private artifact links, or raw evaluator logs. The workbook identity is recorded by filename and SHA-256. No paper result depends on fetching private paths embedded in the source package.

To run a new native evaluation, use the pinned benchmark's installation and adapter instructions, obtain the chosen public task, start and seed the fixture applications, configure the judge, and meet the product's account and recording prerequisites. Preserve the new task, product, access, and judge versions. This is a new evaluation, not a reconstruction of the historical pilot.

\section{Selection and Missing Evidence}
\label{app:selection}
The pilot selection record distinguishes selected numeric scores, selected NA outcomes, untested cells, earlier excluded Brackett attempts, and explicitly excluded Codex attempts. Excluded rows are not silently added to the principal comparison. Treating an excluded incomplete attempt as zero is a sensitivity convention, not evidence that its comprehension was graded as zero.

The recorded exclusions comprise 10 earlier Brackett attempts and 6 Codex attempts. Of the latter, one has an all-answers-empty annotation and five have recording-quality annotations. Keeping the 28 shared cases fixed and including just the all-answers-empty Codex attempt as zero changes its mean from 55.4\% to 54.5\%. This is a sensitivity to one selection decision, not a corrected estimate of performance. Assigning zeros to recording failures would also conflate missing teaching evidence with measured comprehension. Neither convention resolves earlier exclusions or unlisted retries.

A prospective comparison must freeze which repeats enter the result and report failures by stage. It should retain raw answers, item grades, question revisions, judge configuration, capture evidence, allowed learned state, interventions, and the stopping rule. Related takes and variants should remain grouped in any uncertainty analysis. The historical workbook cannot supply missing records merely by being converted to a machine-readable format.

\end{document}